# Handwritten digit Recognition using Support Vector Machine

Anshuman Sharma
(anshuman515@gmail.com)

**Abstract**: Handwritten Numeral recognition plays a vital role in postal automation services especially in countries like India where multiple languages and scripts are used Discrete Hidden Markov Model (HMM) and hybrid of Neural Network (NN) and HMM are popular methods in handwritten word recognition system. The hybrid system gives better recognition result due to better discrimination capability of the NN. A major problem in handwriting recognition is the huge variability and distortions of patterns. Elastic models based on local observations and dynamic programming such HMM are not efficient to absorb this variability. But their vision is local. But they cannot face to length variability and they are very sensitive to distortions. Then the SVM is used to estimate global correlations and classify the pattern. Support Vector Machine (SVM) is an alternative to NN. In Handwritten recognition, SVM gives a better recognition result. The aim of this paper is to develop an approach which improve the efficiency of handwritten recognition using artificial neural network

## 1. Introduction

Handwritten Recognition refers to the process of translating images of hand-written, typewritten, or printed digits into a format understood by user for the purpose of editing, indexing/searching, and a reduction in storage size. Handwritten recognition system is having its own importance and it is adoptable in various fields such as online handwriting recognition on computer tablets, recognize zip codes on mail for postal mail sorting, processing bank check amounts, numeric entries in forms filled up by hand and so on. There are two distinct handwriting recognition domains; online and offline, which are differentiated by the nature of their input signals. In offline system, static representation of a digitized document is used in applications such as cheque, form, mail or document processing. On the other hand, online handwriting recognition (OHR) systems rely on information acquired during the production of the handwriting. They require specific equipment that allows the capture of the trajectory of the writing tool. Mobile communication systems such as Personal Digital Assistant (PDA), electronic pad and smart-phone have online handwriting recognition interface integrated in them. Therefore, it is important to further improve on the recognition performances for these applications while trying to constrain space for parameter storage and improving processing speed. Figure 1 shows an online handwritten Word recognition system. Many current systems use Discrete Hidden Markov Model based recognizer or a hybrid of Neural Network (NN) and Hidden Markov Model (HMM) for the recognition

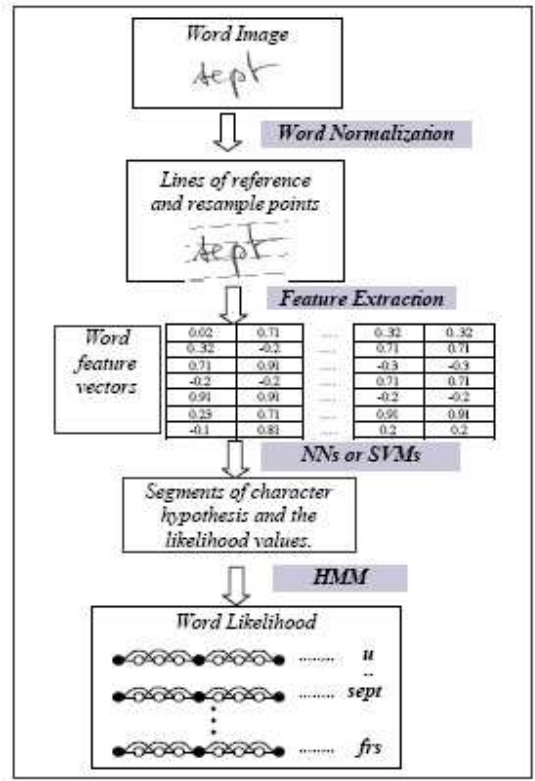

Figure 1- Online word recognition system

Online information captured by the input device first needs to go through some filtration, preprocessing and normalization processes. After normalization, the writing is usually segmented into basic units (normally character or part of character) and each segment is classified and labeled. Using HMM search algorithm in the context of a language model, the most likely word path is then returned to the user as the intended string [1]. Segmentation process can be performed in various ways. However, observation probability for each segment is normally obtained by using a neural network (NN) and a Hidden Markov Model (HMM) estimates the probabilities of transitions within a resulting word path. This research aims to investigate the usage of support vector machines (SVM) in place of NN in a hybrid SVM/HMM recognition system. The main objective is to further improve the recognition rate[6,7] by using support vector machine (SVM) at the segment classification level. This is motivated by successful earlier work by Ganapathiraju in a hybrid SVM/HMM speech recognition (SR) system and the work by Bahlmann [8] in OHR. Ganapathiraju obtained better recognition rate Compared to hybrid NN/HMM SR system. In this work, SVM is first developed and used to tra zin an OHR system using character databases. SVM with probabilistic output are then developed for use in the hybrid system. Eventually, the SVM will be integrated with the HMM module for word recognition. Preliminary results of using SVM for character recognition are given and compared with results using NN reported by Poisson [9]. The following databases were used: IRONOFF, UNIPEN and the mixture IRONOFF-UNIPEN databases.

## 2. Existing Techniques
### 2.1 Modified discrimination function (MQDF) Classifier
G. S. Lehal and Nivedan Bhatt [10] designed a recognition system for handwritten Devangari Numeral using Modified discrimination function (MQDF) classifier. A recognition rate and a confusion rate were obtained as 89% and 4.5% respectively.

### 2.2 Neural Network on Devenagari Numerals
R. Bajaj, L. Dey, S. Chaudhari [11] used neural network based classification scheme. Numerals were represented by feature vectors of three types. Three different neural classifiers had been used for classification of these numerals. Finally, the outputs of
the three classifiers were combined using a connectionist scheme. A 3-layer MLP was used for implementing the classifier for segment-based features. Their work produced recognition rate of 89.68%.

### 2.3 Gaussian Distribution Function
R. J. Ramteke et.al applied classifiers on 2000 numerals images obtained from different individuals of different professions. The results of PCA, correlation coefficient and perturbed moments are an experimental success as compared to MIs. This research produced 92.28% recognition rate by considering 77 feature dimensions.

### 2.4 Fuzzy classifier on Hindi Numerals
M. Hanmandlu, A.V. Nath, A.C. Mishra and V.K. Madasu used fuzzy membership function for recognition of Handwritten Hindi Numerals and produce 96% recognition rate. To recognize the unknown numeral set, an exponential variant of fuzzy membership function was selected and it was constructed using the normalized vector distance.

### 2.5 Multilayer Perceptron
Ujjwal Bhattacharya, B. B. Chaudhuri [11] used a distinct MLP classifier. They worked on Devanagari, Bengali and English handwritten numerals. A back propagation (BP) algorithm was used for training the MLP classifiers. It provided 99.27% and 99.04% recognition accuracies on the original training and test sets of Devanagari numeral database, respectively.

### 2.6 Quadratic classifier for Devanagari Numerals
U. Pal, T. Wakabayashi, N. Sharma and F. Kimura [14] developed a modified quadratic classifier for recognition of offline handwritten numerals of six popular Indian scripts; viz. They had used 64 dimensional features for high-speed recognition. A five-fold cross validation technique has been used for result computation and obtained 99.56% accuracy from Devnagari scripts, respectively.

## 3. Proposed Approach
### 3.1 Support Vector Machine (SVM)
SVM in its basic form implement two class classifications. It has been used in recent years as an alternative to popular methods such as neural network. The advantage of SVM, is that it takes into account both experimental data and structural behavior for better generalization capability based on the principle of structural risk minimization (SRM). Its formulation approximates SRM principle by maximizing the margin
of class separation, the reason for it to be known also as large margin classifier. The basic SVM formulation is for linearly separable datasets. It can be used for nonlinear datasets by indirectly mapping the nonlinear inputs into to linear feature space where the maximum Margin decision function is approximated. The mapping is done by using a kernel function. Multi class classification can be performed by modifying the 2 class scheme. The objective of recognition is to interpret a sequence of numerals taken from the test set. The architecture of proposed system is given in fig. 3.The SVM (binary classifier) is applied to multi class numeral recognition problem by using one-versus-rest type method. The SVM is trained with the training samples using linear kernel. Classifier performs its function in two phases; Training and Testing. [29] After preprocessing and Feature Extraction process, Training is performed by considering the feature vectors which are stored in the form of matrices. Result of training is used for testing the numerals. Algorithm for Training is given in algorithm.

**3.2 Statistical Learning Theory**
Support Vector Machines have been developed by Vapnik in the framework of Statistical Learning Theory [13]. In statistical learning theory (SLT), the problem of classification in supervised learning is formulated as follows:
We are given a set of *l* training data and its class, $\{(x1,y1)...(xl,yl)\}$ in $Rn \times R$ sampled according to unknown joint probability distribution $P(x,y)$ characterizing how the classes are spread in $Rn \times R$. To measure the performance of the classifier, a loss function $L(y,f(\mathbf{x}))$ is defined as follows:

$$L(y, f(x)) = \begin{cases} 0 \text{ if } y = f(x) \\ 1 \text{ if } y \neq f(x) \end{cases}$$

$L(y,f(\mathbf{x}))$ is zero if *f* classifies x correctly, one otherwise. On average, how *f* performs can be described by the Risk functional:

$$R(f) = \int L(y, f(x)) dP(x,y)$$

ERM principle states that given the training set and a set of possible classifiers in the hypothesis space *F*, we Should choose $f \subset F$ that minimizes *Remp(f)*. However, which generalizes well to unseen data due to over fitting phenomena. *Remp(f)* is a poor, over-optimistic approximation of *R(f)*, the true risk. Neural network classifier relies on ERM principle. The normal practice to get a more realistic estimate of generalization error, as in neural network is to divide the available data into training and test set. Training set is used to find a Classifier with minimal empirical error (optimize the weight of an MLP neural networks) while the test set is used to find the generalization error (error rate on the Test set). If we have different sets of classifier hypothesis space *F1, F2 ... e.g.* MLP neural networks with different topologies, we can select a classifier from each hypothesis space (each topology) with minimal *Remp(f)* and choose the final classifier with minimal generalization error. However, to do that requires designing and training potentially large number of individual classifiers. Using SLT, we do not need to do that. Generalization error can be directly minimized by minimizing an upper bound of the risk functional *R(f)*.
The bound given below holds for any distribution P(x,y) with probability of at least 1- η :

$$R(f) \leq R_{emp}(f) + \phi(\frac{h}{l}, \frac{log(\eta)}{l})$$

where the parameter h denotes the so called VC (Vapnik-Chervonenkis) dimension. $\phi$ is the confidence term defined by Vapnik [10] as :

$$\phi(\frac{h}{l}, \frac{log(\eta)}{l}) = \sqrt{\frac{h(log\frac{2l}{h}+1) - log(\frac{\eta}{4})}{l}}$$

ERM is not sufficient to find good classifier because even with small *Remp(f),* when **h** is large compared to *l*, $\phi$
will be large, so *R(f)* will also be large, ie: not optimal. We actually need to minimize *Remp(f)*and $\phi$ at the same time, a process which is called structural risk
Minimization (SRM). By SRM, we do not need test set for model selection anymore. Taking different sets of classifiers *F1, F2 ...* with known *h1, h2 ...* we can select *f*
from one of the set with minimal *Remp(f)*, compute $\phi(\frac{h}{l}, \frac{log(\eta)}{l})$
 and choose a classifier with minimal *R(f)*.No more evaluation on test set needed, at least in theory. However, we still have to train potentially very large
number of individual classifiers. To avoid this, we want to make **h** tunable (ie: to cascade a potential classifier *Fi* with VC dimension = h and choose an optimal *f* from an optimal *Fi* in a single optimization step. This is done in large margin classification.

**3.3 SVM formulations**
SVM is realized from the above SLT framework. The simplest formulation of SVM is linear, where the decision hyper plane lies in the space of the input data **x**.

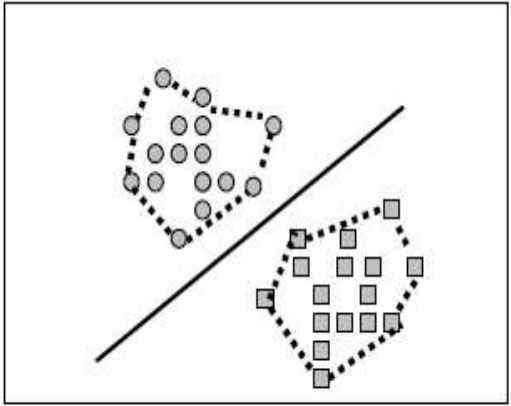

Figure 2-Optimal plane bisects closest points in convex hulls

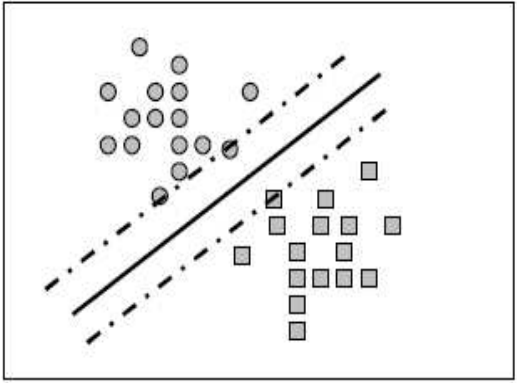

Figure 3- Optimal plane maximize margin

In this case the hypothesis space is a subset of all hyper planes of the form: *f(x) = w·x +b.* SVM finds an optimal hyper plane as the solution to the learning Problem which is geometrically the furthest from both classes since that will generalize best for future unseen data.

There are two ways of finding the optimal decision hyper plane. The first is by finding a plane that bisects the two closest points of the two convex hulls defined by the set of points of each class, as shown in figure 2. The second is by maximizing the margin between two supporting planes as shown in figure 3.

Both methods will produce the same optimal decision plane and the same set of points that support the solution (the closest points on the two convex hulls in figure 2 or the points on the two parallel supporting planes in figure 3). These are called the support vectors.

## 4. Feature Extraction
### 4.1 Moment Invariants
The moment invariants (MIs) [1] are used to evaluate seven distributed parameters of a numeral image. In any character Recognition system, the characters are processed to extract features that uniquely represent properties of the character. Based on normalized central moments, a set of seven moment invariants is derived. Further, the resultant image was thinned and seven moments were extracted. Thus we had 14 features (7 original and 7 thinned), which are applied as features for recognition using Gaussian Distribution Function. To increase the success rate, the new features need to be extracted by applying Affine Invariant Moment method.

### 4.2 Affine Moment Invariants
The Affine Moment Invariants were derived by means of the theory of algebraic invariants. Full derivation and comprehensive discussion on the properties of invariants can be found. Four features can be computed for character recognition. Thus overall 18 features have been used for Support Vector Machine.

## 5. Experiment
### 5.1 Data Set Description
In this paper, the UCI Machine learning data set are used. The UCI Machine Learning Repository is a collection of databases, domain theories, and data generators that are used by the machine learning community for the empirical analysis of machine learning algorithms. One of the available datasets is the Optical Recognition of the Handwritten Digits Data Set. The dataset of handwritten assamese characters by collecting samples from 45 writers is created. Each writer contributed 52 basic characters, 10 numerals and 121 assamese conjunct consonants. The total number of entries corresponding to each writer is 183 (= 52 characters + 10 numerals + 121 conjunct consonants). The total number of samples in the dataset is 8235 ( = 45 × 183 ).

The handwriting samples were collected on an iball 8060U external digitizing tablet connected to a laptop using its cordless digital stylus pen. The distribution of the dataset consists of 45 folders. This file contains information about the character id (ID), character name (Label) and actual shape of the character (Char).

In the raw Optdigits data, digits are represented as 32x32 matrices. They are also available in a pre-processed form in which digits have been divided into non-overlapping blocks of 4x4 and the number of on pixels have been counted in each block. This generated 8x8 input matrices where each element is an integer in the range 0.16.

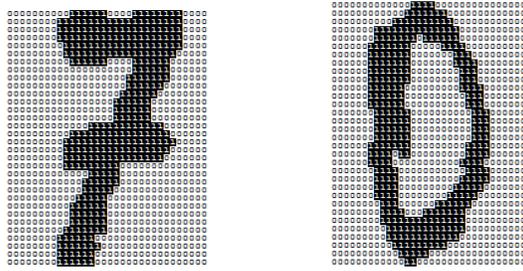

**Fig4** - Sample digits extracted from the raw Optdigits dataset.

### 5.2 Data Preprocessing

For the experiments using SVM, example isolated characters are preprocessed and 7 local features for each point of the spatially resample online signal were extracted. For each example character there are 350 feature values as input to the SVM. We use SVM with RBF kernel, since RBF kernel has been shown to generally give better recognition result [7]. Grid search was done to find the best values for the C and gamma (representing $\frac{1}{2\sigma^2}$ in the original RBF kernel formulation) for the final SVM models and by that, C = 8 and gamma = $2^{-5}$ were chosen.

### 5.4 Experimental Results

#### 5.4.1 *Test application Analysis*

The test application accompanying the source code can perform the recognition of handwritten digits. To do so, open the application (preferably outside Visual Studio, for better performance). Click on the menu File and select Open. This will load some entries from the Optdigits dataset into the application.

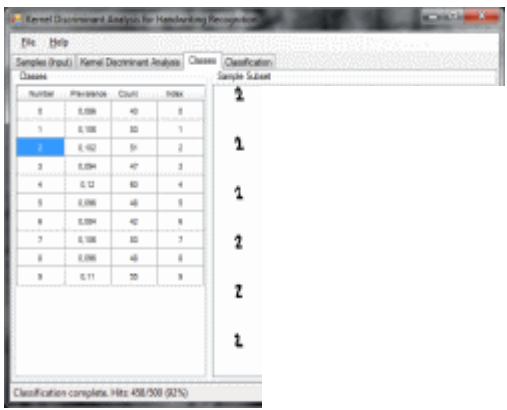

*Fig.5: Optdigits da... application*

To perform the analysis, click the Run Analysis button. Please be aware that it may take some time. After the analysis is complete, the other tabs in the sample application will be populated with the analysis' information. The level of importance of each factor found during the discriminant analysis is plotted in a pie graph for easy visual inspection.

Once the analysis is complete, we can test its classification ability in the testing data set. The green rows have been correctly identified by the discriminant space Euclidean distance classifier. We can see that it correctly identifies 98% of the testing data. The testing and training data set are disjoint and independent.

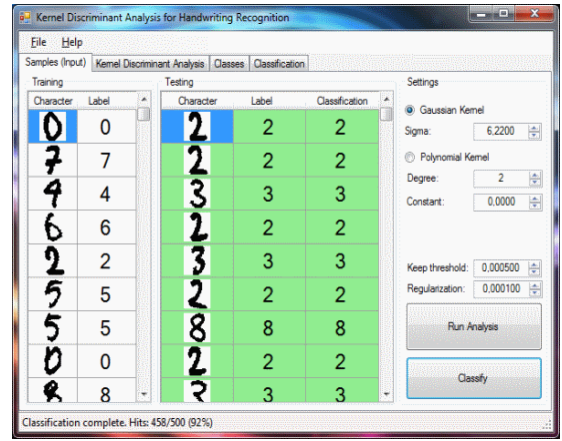

*Fig.6: Using the default values in the application*

### 5.5 Results

After the analysis has been completed and validated, we can use it to classify the new digits drawn directly in the application. The bars on the right show the relative response for each of the discriminant functions. Each class has a discriminant function that outputs a closeness measure for the input point. The classification is based on which function produces the maximum output.

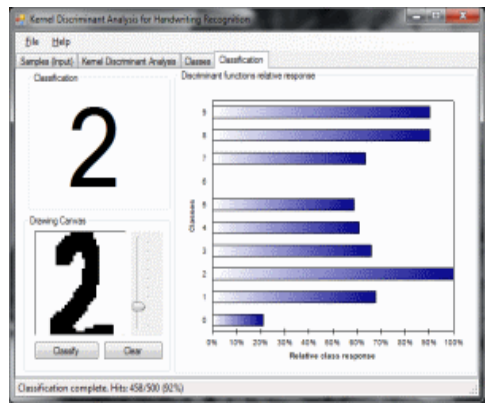

*Fig 6: We can see the analysis also performs rather well on completely new and previously unseen data.*

Experiments were performed on different samples having mixed scripting languages on numerals using single hidden layer.

| Data Set | Training set size | Testing Set Size | Training Set Accuracy | Test Set accuracy | Number of Support Vector | Approx training time |
|---|---|---|---|---|---|---|
| Digit | 1778 | 6270 | 96 | 97 | 3014 | 497 |

**Table 1**: Detail Recognition performance of SVM on UCI datasets

| Data Set | Algorithm | Total Sample | Recognized Sample | Accuracy |
|---|---|---|---|---|
| Optical data set | Hidden Markov Model | 376 | 345 | 92 |
| Optical data set | Support Vector Machine | 376 | 366 | 96 |

**Table 2:** Detail Recognition performance of SVM and HMM on UCI datasets

| Class | No. of Example in data set | Recognition Rate In SVM | Recognition Rate in HMM |
|---|---|---|---|
| 1 | 389 | 99% | 96% |
| 2 | 380 | 98% | 94% |
| 3 | 389 | 100% | 96% |
| 4 | 387 | 97% | 94% |
| 5 | 376 | 99% | 93% |
| 6 | 377 | 97% | 92% |
| 7 | 387 | 100% | 95% |
| 8 | 380 | 96% | 93% |
| 9 | 382 | 96% | 92% |

**Table 3:** Recognition Rate of Each Numeral in DATASET.

It is observed that recognition rate using SVM is higher than Hidden Markov Model. However, free parameter storage for SVM model is significantly higher. The memory space required for SVM will be the number of support vectors multiply by the number of feature values (in this case 350). This is significantly large compared to HMM which only need to store the weight. HMM needs less space due to the weight-sharing scheme. However, in SVM, space saving can be achieved by storing only the original online signals and the penup/ pen-down status in a compact manner. During recognition, the model will be expanded dynamically as required. Table 3 shows the comparison of recognition rates between HMM and SVM using all three databases. SVM clearly outperforms in all three isolated character cases.

The result for the isolated character cases above indicates that the recognition rate for the hybrid word recognizer could be improved by using SVM instead of HMM. Thus, we are currently implementing word recognizer using both HMM and SVM and comparing their performance.

**Fig 7:** Graph Representation between HMM and SVM

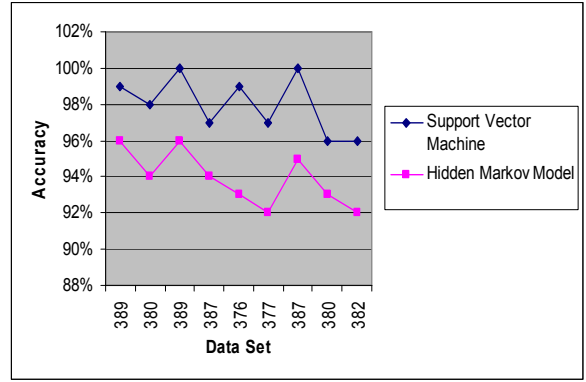

## 6. Conclusion

Handwriting recognition is a challenging field in the machine learning and this work identifies Support Vector Machines as a potential solution. The number of support vectors can be reduced by selecting better C and gamma parameter values through a finer grid search and by reduced set selection Work on integrating the SVM character recognition framework into the HMM based word recognition framework is on the way. In the hybrid system, word preprocessing and normalization needs to be done before SVM is then used for character hypothesis recognition and word likelihood computation using HMM. It is envisaged that, due to SVM's better discrimination capability, word recognition rate will be better than in a HMM hybrid system.